\documentclass[conference]{IEEEtran}

\usepackage{amsmath,amssymb,amsfonts}
\usepackage{bbm}
\usepackage{booktabs}
\usepackage{bbding}
\usepackage{cite}
\usepackage{float}
\usepackage[T1]{fontenc}
\usepackage{graphicx}
\usepackage{svg}
\usepackage[utf8]{inputenc}
\usepackage{makecell}
\usepackage{mathtools}
\usepackage{multicol}
\usepackage{multirow}
\usepackage{microtype}
\usepackage{newtxmath}
\usepackage{pgfplots}
\usepackage{stfloats}
\usepackage{subcaption}
\usepackage{tablefootnote}
\usepackage{textcomp}
\usepackage{wrapfig}
\usepackage{xcolor}

\pgfplotsset{compat=1.18}

\begin{document}

\title{Exploring the Impact of Temperature Scaling in Softmax for Classification and Adversarial Robustness}

\author{\IEEEauthorblockN{1\textsuperscript{st} Hao Xuan}
\IEEEauthorblockA{\textit{Electrical and Computer Engineering} \\
\textit{University of Alberta}\\
Edmonton, Canada \\
hxuan@ualberta.ca}
\and
\IEEEauthorblockN{2\textsuperscript{nd} Bokai Yang}
\IEEEauthorblockA{\textit{Electrical and Computer Engineering} \\
\textit{University of Alberta}\\
Edmonton, Canada \\
bokai5@ualberta.ca }
\and
\IEEEauthorblockN{3\textsuperscript{rd} Xingyu Li}
\IEEEauthorblockA{\textit{Electrical and Computer Engineering} \\
\textit{University of Alberta}\\
Edmonton, Canada \\
xingyu@ualberta.ca }}

\maketitle
\begin{abstract}
The softmax function is a fundamental component in deep learning.  This study delves into the often-overlooked parameter within the softmax function, known as "temperature," providing novel insights into the practical and theoretical aspects of temperature scaling for image classification. Our empirical studies, adopting convolutional neural networks and transformers on multiple benchmark datasets, reveal that moderate temperatures generally introduce better overall performance. Through extensive experiments and rigorous theoretical analysis, we explore the role of temperature scaling in model training and unveil that temperature not only influences learning step size but also shapes the model's optimization direction. Moreover, for the first time, we discover a surprising benefit of elevated temperatures: enhanced model robustness against common corruption, natural perturbation, and non-targeted adversarial attacks like Projected Gradient Descent. We extend our discoveries to adversarial training, demonstrating that, compared to the standard softmax function with the default temperature value, higher temperatures have the potential to enhance adversarial training. The insights of this work open new avenues for improving model performance and security in deep learning applications. 

\end{abstract}

\section{Introduction}
Deep learning has achieved dramatic breakthroughs in recent years, excelling in tasks such as image classification~\cite{NIPS2012_c399862d}, nature language processing (NLP)~\cite{DBLP_abs-1810-04805}, and semantic segmentation~\cite{Zhao_2018_ECCV}. A critical component of most deep learning methods is the softmax function, which normalizes a set of real values into probabilities. The generalized softmax function incorporates a parameter known as "temperature," which controls the softness of the output distribution. Despite its importance in theory, the impact of temperature scaling on classification tasks has been relatively underexplored, particularly in contrast to its use in other areas such as knowledge distillation~\cite{hinton2015distilling}, contrastive learning~\cite{Wang_2021_CVPR}, confidence calibration~\cite{pereyra2017regularizing}, and natural language processing.
Specifically, though the temperature scaling has occasionally been applied in prior experimentation~\cite{shafahi2019label,engstrom2018evaluating,kanai2021constraining}, these studies often integrate additional complex techniques such as Gaussian noise injection in~\cite{shafahi2019label}, adversarial training in~\cite{engstrom2018evaluating,prach2022almost}, and innovative quadratic activation functions in~\cite{kanai2021constraining}, making it challenging to isolate and understand the specific contribution of temperature scaling to the overall system performance. Consequently, the specific role of temperature in classification tasks remains ambiguous. Previous study by~\cite{DBLP_abs-2010-07344} has hinted at the potential benefits of temperature scaling, but a comprehensive investigation is still lacking.

This study aims to fill this gap by conducting extensive experiments to explore the practical and theoretical aspects of temperature scaling in the softmax function for image classification. We employ convolutional neural networks (CNNs) and transformers on multiple benchmark datasets, including CIFAR-10~\cite{krizhevsky2009learning}, CIFAR-100~\cite{krizhevsky2009learning}, and Tiny-ImageNet~\cite{le2015tiny}, to systematically analyze the effects of different temperature values. Our empirical results consistently show that moderate temperatures generally improve overall performance, challenging the conventional knowledge derived from contrastive learning that low temperature facilitates representation learning.

We also delve into the theoretical underpinnings of temperature scaling in model training. Our analysis reveals that temperature not only influences the learning step size but also shapes the model’s optimization direction. Specifically, lower temperatures focus the model's learning on error-prone classes, while higher temperatures promote a more balanced learning across all classes. This insight is crucial for understanding the nuanced effects of temperature scaling on model optimization.

Furthermore, we uncover a surprising benefit of elevated temperatures: enhanced model robustness against common corruptions, natural perturbations, and non-targeted adversarial attacks, such as Projected Gradient Descent (PGD). We extend our investigation to adversarial training introduced by~\cite{madry2018at}, demonstrating that higher temperatures can potentially enhance the robustness of models trained with adversarial methods compared to those using the standard softmax function with the default temperature.

In summary, this work provides new perspectives on the practical applications and theoretical implications of temperature scaling in the softmax function. Our contributions can be summarized as follows:
\begin{itemize}
\item We conduct extensive experiments demonstrating that applying a reasonably large temperature during model training improves overall performance.
\item We discover that models trained with elevated temperatures exhibit enhanced robustness against gradient-based untargeted adversarial attacks.
\item Additionally, we show the potential of integrating temperature control into adversarial training to boost model performance and security in deep learning applications.
\end{itemize}

\section{Related Works}
The softmax function has been a longstanding component of neural networks, usually used to normalize a vector of real values into probabilities. Modulating the temperature scaling factor within the softmax function allows for reshaping the probability distribution. This section provides a concise overview of the application of temperature scaling in various computational tasks. 

\textbf{Knowledge Distillation} proposed by \cite{hinton2015distilling} is one innovative way to transfer knowledge from a teacher model to a student model.
Temperature is utilized during training to control both the student and teacher model's output. The author argues that lower temperatures make the distillation assign less weight to logits that are much smaller than the average. Conversely, employing larger temperatures softens the probability distribution and pays more attention to the unimportant part of the logit. Larger temperatures are proven to be beneficial in the distillation process since the hard-target term already ensures the dominant part of the logit (target class) is correct. By focusing on the remaining logit, the student model can capture more fine-grained information from the teacher model. Note that despite various temperatures used during training, it is set to 1 when the model is deployed.

\textbf{Model Confidence Calibration} usually utilizes temperature scaling to address the over-confident issue in deep learning~\cite{pmlr-v70-guo17a, NIPS2017_9ef2ed4b, NEURIPS2021_8420d359}. It centers on estimating predictive uncertainty to match its expected accuracy~\cite{pmlr-v80-kumar18a, NEURIPS2019_8ca01ea9}. Despite multiple generic calibration methods being proposed, temperature scaling proposed by~\cite{pmlr-v70-guo17a} remains a baseline method for being simple, effective and able to apply to various cases without major expense. The motivation behind temperature scaling is simple, since the goal is to control the network's confidence to match its accuracy, applying temperature to the softmax function that can directly modify the probability distribution seems a perfect fit for the problem. During training, a validation set is needed to find the ideal temperature parameter for the network, and the same temperature is used when deployed.

\textbf{Contrastive Learning} is one paradigm for unsupervised learning~\cite{DBLP:abs-1807-03748, Wu_2018_CVPR}. To achieve a powerful feature encoder, it utilizes contrastive loss to pull similar samples close and push negative pairs away in the latent space. 
Although the temperature has long existed as a hyper-parameter in contrastive loss, its actual mechanism is just understudied recently. 
~\cite{Wang_2021_CVPR} analyze the contrastive loss closely and find that as the temperature decreases, the distribution of the contrastive loss becomes sharper, which applies larger penalties to samples similar to the anchor data. Also, uniformity of feature distribution increases, indicating the embedding feature distribution aligns with a uniform distribution better \cite{pmlr-v119-wang20k}. 

\textbf{Temperature Scaling in Image Classification} has occasionally been utilized in the experimental sections of prior studies, yet focused investigations on this subject remain limited. For example, previous studies aiming to improve adversarial robustness have utilized temperature scaling to adjust logits within their experimentation~\cite{shafahi2019label,engstrom2018evaluating,kanai2021constraining}. However, these studies often integrate additional complex techniques such as Gaussian noise injection~\cite{shafahi2019label}, adversarial training~\cite{engstrom2018evaluating,prach2022almost}, and innovative quadratic activation functions~\cite{kanai2021constraining}, making it challenging to isolate and understand the specific contribution of temperature scaling to the overall system performance. In contrast, our study narrows its focus to investigating the direct impact of temperature scaling applied through the softmax function on model optimization processes. Among the few related works, "The Temperature Check" by~\cite{DBLP_abs-2010-07344} is notably relevant to our discussion. It mainly explores the dynamics of model training by considering factors such as temperature, learning rate, and time, and presents an empirical finding that a model's generalization performance is significantly influenced by temperature settings. While our observations align with these findings, our research approaches the issue from a different perspective of gradient analysis. Specifically, we delve into how temperature scaling impacts model optimization process. Furthermore, our study broadens the scope of inquiry by assessing the effect of temperature scaling on a model's resilience to common corruptions and adversarial attacks, thereby adding a new dimension to the existing research. 

\section{Preliminary}

\subsection{Softmax Function}

Given a set of real numbers, $X=\{x_1,..,x_N\}$, the generalized softmax function can be used to normalize $X$ into a probability distribution.
\begin{equation}
    \mathbb{S}(X) = \frac{\exp(X/\tau)}{\sum_i \exp(x_i/\tau)},
\end{equation}
where $\mathbb{S}$ represents the softmax function and $\tau$ is the temperature scaling factor. The temperature $\tau$ controls the smoothness (softness) of the probability it produces. Specifically, when $\tau \rightarrow \infty$, the output tends toward a uniform distribution; while when $\tau=0$, the softmax function assigns a probability of 1 to the element with the highest value and a probability of 0 to the rest. The standard (unit) softmax function, with $\tau=1$, is widely used in conventional classification tasks. 

\subsection{Problem Definition and Notation}

We consider multi-category classification in this study, where paired training data $\{\mathscr{X},\mathscr{Y}\}=\{(x,y)|x\in \mathbb{R}^{H\times L \times N}, y\in \mathbb{R}^{1\times M}\}$ are drawn from a data distribution $\mathcal{D}$. Here, $H, L, N$ are the dimension of a sample $x$, $M$ is the number of categories, and $y$ is a one-hot vector indicating the class of the input $x$. A classifier, $\mathcal{C}:\mathscr{X}\xrightarrow{}\mathscr{Y}$, is a function predicting the label $y$ for a given data $x$. That is $C(x)=y$. In the canonical classification setting, a neural network classifier, $\mathcal{C}=(f,W)$, is usually composed of a feature extractor $f$ parameterized by $\theta$ and a weight matrix $W$. $f$ is a function mapping the input $x$ to a real-valued vector $f(x)$ in the model’s penultimate layer and $W=(w_1,...,w_M)$ represents the coefficients of the last linear layer before the softmax layer. So the likelihood probability of data $x$ corresponding to the $M$ categories can be formulated as 
\begin{equation}
    \hat{y}=\mathcal{C}(x)=\mathbb{S}(W^T f(x)).
    \label{eqn:classification}
\end{equation}
Note that each vector $w_i$ in matrix $W$ can be considered as the prototype of class $i$ and the production $W^T f(x)$ in Eqn.~\ref{eqn:classification} quantifies the similarity between the feature $f(x)$ and different class-prototypes.

During training, the model $\mathcal{C}=(f,W)$ is optimized to minimize a specific loss, usually a Cross-Entropy (CE) loss.
\begin{equation}\label{eq:ce}
    \begin{split}
     L_{ce}(x)=-y\log\hat{y}
      = -\log \left[ \frac{\exp(w_i^T\cdot f(x)/\tau)}{\sum_{j=1}^N \exp(w_j^T\cdot f(x)/\tau)} \right]
    \end{split}
\end{equation}
Though $\tau=1$ is the default setting in classification tasks, we preserve $\tau$ in the Eqn.s to facilitate theoretical analysis.

\section{Gradient Analysis}
To investigate the impact of temperature scaling factors for model optimization in classification tasks, we calculate the loss gradients with respect to the training parameters in the model. 
Specifically, given a data sample $x$ from the $i^{th}$ category, we refer to $w_i$ as the positive class prototype and the rest, $w_j$ for $j\neq i$, as the negative class prototypes. Then the gradients with respect to the positive class prototype, negative class prototypes, and the encoder are:
\begin{equation} \label{Eq:wi}
\begin{split}
    \frac{\partial L_{ce}(x)}{\partial w_i}=\frac{1}{\tau}[\mathbb{S}(w_i^T\cdot f(x)/\tau)-1]f(x)=\frac{1}{\tau}[P_i^{\tau}(x)-1]f(x),
\end{split}
\end{equation}
\begin{equation} \label{Eq:wj}
\begin{split}
    \frac{\partial L_{ce}(x)}{\partial w_j}= \frac{1}{\tau}\mathbb{S}(w_j^T\cdot f(x)/\tau)f(x)= \frac{1}{\tau} P_j^{\tau}(x)f(x),
\end{split}
\end{equation}
\begin{equation} \label{Eq:f}
\begin{split}
    \frac{\partial L_{ce}(x)}{\partial f}
    =\frac{1}{\tau}\left[\sum\limits_{j\neq i}w_k P_j^{\tau}(x)-w_i[1-P_i^{\tau}(x)]\right].
\end{split}
\end{equation}

\textbf{Learning rate:} In Eqn.~\ref{Eq:wi}, \ref{Eq:wj}, \ref{Eq:f}, since $0<P_j^{\tau}(x)<1$, the actual learning rate is inversely proportional to the temperature $\tau$. That is, larger temperatures lead to a reduced gradient step in model update, while smaller temperatures not only increase the gradient step. Furthermore, when the sample $x$ is misclassified, smaller temperatures give a further boost on updating $w_i$ and $w_j$ for $j=argmax(P_j^{\tau}(x)f(x))$, because smaller temperatures in softmax function lead to shaper distributions. 

\begin{figure}[t]
    \centering
    \includegraphics[width=0.48\textwidth]{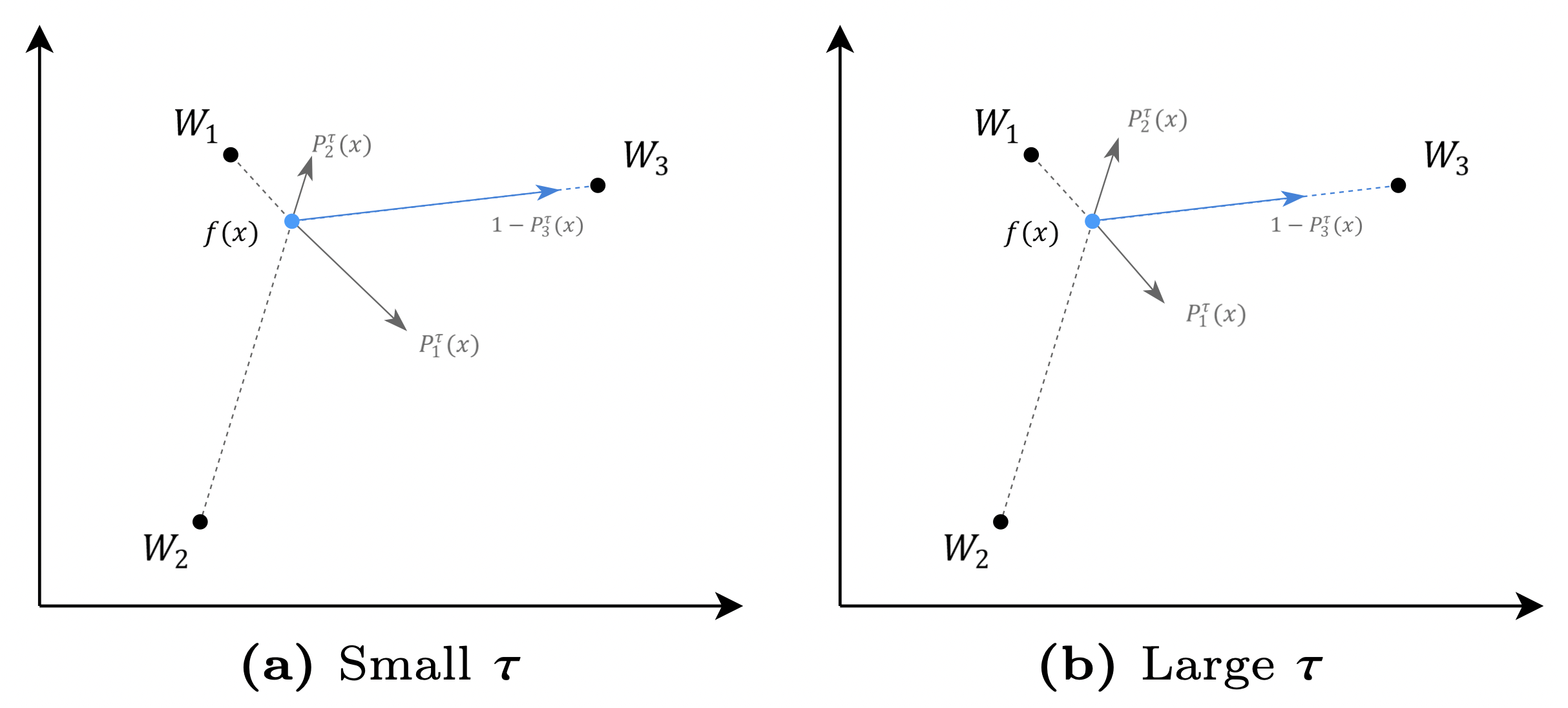}
    \caption{Demonstration of the model optimization direction with different temperatures. $f(x)$ is the latent code of a data sample from category 3. Since $f(x)$ is close to the negative class prototype $w_1$, the CE loss with respect to the encoder $f$ yields a large gradient toward the groundtruth $w_3$. However, with different temperature factors, the gradients associated with the negative classes are different: low temperature makes the update more biased by the hard class (a), while an elevated temperature leads to more equalized gradients (b).}
    \label{class_prototype}
\end{figure}

\textbf{Optimization direction:} From Eqn.~\ref{Eq:wi}, the positive class prototype $w_i$ is updated toward $f(x)$ in the latent space. In contrast, the negative prototypes $w_j$ move away from the direction of $f(x)$ according to Eqn.~\ref{Eq:wj}. The optimization direction of $f(x)$ is a weighted sum of all class prototypes, as shown in Eqn.~\ref{Eq:f}. The fundamental optimization policy is to update the trainable parameters of the encoder in such a way that $f(x)$ moves closer to the positive class prototype and farther away from the negative class prototypes in the latent space. However, when we take the temperature parameter into account, we find that temperature has an impact on the update direction of $f(x)$. Specifically, when the temperature is low, the probability distribution produced by the softmax function is sharper, leading to significant differences in probability values among different prototypes. Consequently, the update direction of the encoder $f$ is predominantly influenced by the class prototype with the highest probability and the positive class prototype (if they are different). Fig.~\ref{class_prototype}(a) visualizes the bias toward the hard class in model optimization, where $f(x)$ is the latent code of a data sample from category 3. In contrast, when the temperature is high, the differences in probability values among different prototypes are relatively smaller, and the encoder $f$ updates with a mixture of all class prototype directions, as demonstrated in Fig.~\ref{class_prototype}(b). In other words, a low temperature makes the model focus on learning hard-class pairs, while a high temperature de-biases the influence among different classes for a balanced learning.

Moreover, when considering all the samples in one batch, the compound gradient of all $N$ samples are
\begin{equation} \label{Eq:batch_wi}
\begin{split}
    \sum\limits^N_{n=1}\frac{\partial L_{ce}(x_n)}{\partial w_i}=-\frac{1}{\tau}\sum\limits^N_{n=1}f(x_n)[1-P^{\tau}_i(x_n)],
    \end{split}
\end{equation}
\begin{equation} \label{Eq:batch_wj}
\begin{split}
    &\sum\limits^N_{n=1}\frac{\partial L_{ce}(x_n)}{\partial w_k}=\frac{1}{\tau}\sum\limits^N_{n=1}f(x_n) P^{\tau}_k(x_n),
    \end{split}
\end{equation}
\begin{equation} \label{Eq:batch_f}
\begin{split}
    \sum\limits^N_{n=1}\frac{\partial L_{ce}(x_n)}{\partial f}=\frac{1}{\tau}\sum\limits^N_{n=1}\left[\sum\limits_{k\neq i}w_k P^{\tau}_k(x_n)
    -w_i [1-P^{\tau}_i(x_n)]\right].
\end{split}
\end{equation}
Similar to the single sample case, when optimizing in a whole batch, with small temperatures, the model focuses on learning misclassified samples (i.e. hard samples), 
whereas higher temperatures help de-bias the update direction and distribute similar weight to all samples.

\begin{figure*}[t]
    \centering
    \includegraphics[width=0.98\textwidth]{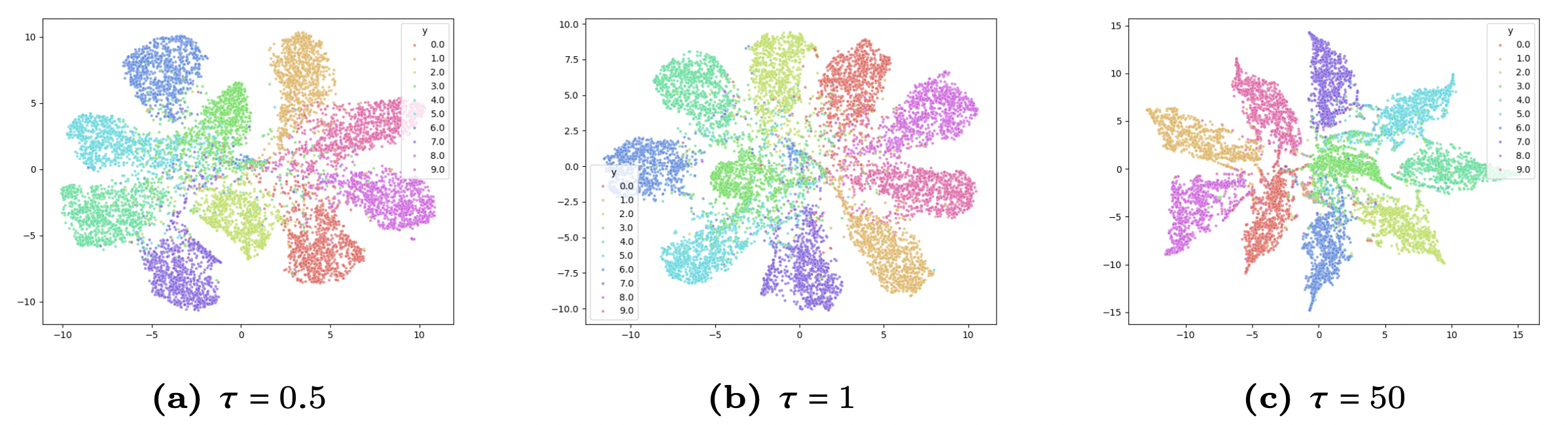}
    \caption{T-SNE~\cite{tsne} visualization of the CIFAR10 sample distribution after the ResNet50 encoder with different temperatures.}
    \label{cluster01}
\end{figure*}

\begin{figure*}[htbp]
    \centering
    \includegraphics[width=0.98\textwidth]{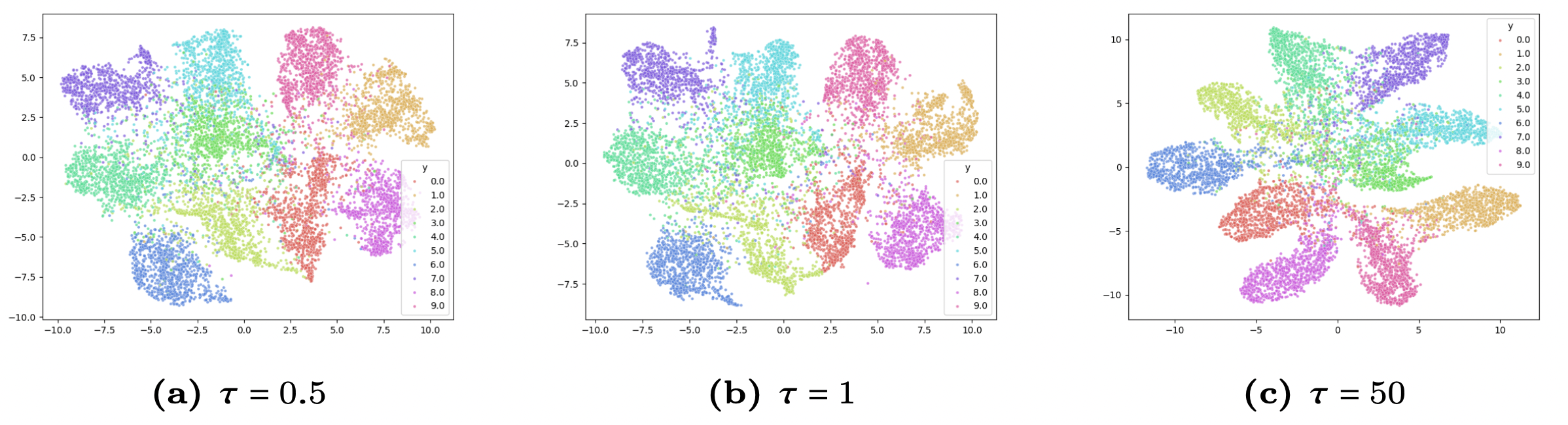}
    \caption{T-SNE~\cite{tsne} visualization of the CIFAR10 sample distribution after the VIT encoder with different temperatures.}
    \label{cluster02}
\end{figure*}

\section{Empirical Analysis and Discussion}
As discussed in Section 4, applying a small temperature encourages a model to learn more about hard (misclassified) samples and hard (error-prone class) classes. A low temperature, however, leads to more equitable learning across different classes and data points. Theoretically, both approaches to optimize feature distribution sound reasonable, with low temperatures focusing on weaker classes and high temperatures decreasing inequality across all negative classes. We argue that which optimization strategy is better for classification tasks remains an empirical problem.  

\subsection{Experiment Setting}
We conduct image classification on multiple benchmarks (i.e. CIFAR10, CIFAR100, and Tiny-ImageNet) and their extended Common Corruptions and Perturbations sets (i.e. CIFAR10-C, CIFAR100-C, and Tiny-ImageNet-C with corruption strentgh being 3) to investigate the impact of temperature scaling. In addition, we also evaluate the model's robustness against adversarial attacks such as PDG20~\cite{madry2018at} and C\&W~\cite{carlini2017towards}. Both attacks are bounded by the $l_\infty$ box with the same maximum perturbation $\epsilon=8/255$.  

To get a comprehensive evaluation, we set $\tau\in \{0.1,0.5,1,10,30,50,70,100\}$. Unless stated otherwise, we takes ResNet50 
and VIT-small-patch16-224 as the CNN and transformer backbones, respectively. The ResNet50 is trained from scratch, with SGD optimizer and learning rate setting to 0.1. We also utilize the Cosine Annealing scheduler to better train the model. The transformer is pretrained on ImageNet-21K and finetuned on the target dataset using Adam optimizer. All experiments run on one RTX3090.

To clarify, the temperature scaling only involves in model training in this study, but not model evaluation and attacks. All empirical evaluation and adversarial sample generation by PGD and C\&W are based on the standard cross entropy, i.e. $\tau=1$. Thus, attack gradients are not attenuated, reflecting model's true sensitivity to data perturbation.

\subsection{Experiment Results}
The quantitative results on CNN and Transformer are summarized in Table~\ref{tb:main_CNN} and Table~\ref{tb:main_VIT}, respectively. For the CNN model, ResNet50, training from scratch, the standard accuracy increases with the temperature increase. Furthermore, CNN models trained at elevated temperatures show more robustness against naturally corrected images. 
We believe that such improvements are majorly attributed to better model optimization with leveraged temperature. For the transformer finetuned on the target set, the standard accuracy and robustness against natural corruptions and perturbations is quite stable. We hypothesize that such stable performance is due to the fact that ViT has already been pre-trained on ImageNet and has reached a relatively high-quality state. Additionally, we observed that the model's adversarial robustness gradually improves with increasing temperature.

Clustering is a crucial metric when measuring how an encoder performs. In classification, a good encoder should be able to gather samples from the same class while separating clusters of different classes. Fig.~\ref{cluster01} and Fig.~\ref{cluster02} present 2D TSNE visualization of the CIFAR10 sample distribution by ResNet50 and transformer. We observe a similar trend: low temperatures lead to more mixed clusters, while models trained with elevated temperatures have better cluster effects. These empirical observations also explain the improved classification performance on clean and non-adversarial perturbations, as well as stronger adversarial robustness, with high temperature in Table~\ref{tb:main_CNN} and Table~\ref{tb:main_VIT}.

\begin{table*}[t]
    \caption{Model performance and Robustness against Common Corruptions and Adversarial attacks (\%) under different temperatures with ResNet50 trained from scratch. -C in the table represents the corresponding Common Corruptions and Perturbations set.}
    \centering
    \renewcommand{\arraystretch}{1.6}
    \setlength{\tabcolsep}{4pt}
    \scalebox{1}{
    \begin{tabular}{lcccccccccccc}
        \toprule
         \multirow{2}[2]{*}{Temp.}
         & \multicolumn{4}{c}{CIFAR10}& \multicolumn{4}{c}{CIFAR100}& \multicolumn{4}{c}{Tiny-Imagenet}\\
        \cmidrule(lr){2-5} \cmidrule(lr){6-9} \cmidrule(lr){10-13} 
         & Clean& -C & PGD20& C\&W& Clean& -C& PGD20& C\&W& Clean& -C &PGD20& C\&W\\
        \hline\hline
        $\tau=0.1$& 90.05& 73.31 & 0& 27.79& 70.39& 44.52 & 0& 14.32& 54.53& 12.63& 0& 23.17\\
        \hline
        $\tau=0.5$& 94.17& 72.51& 0& 16.03& 74.79& 45.41 & 0& 8.44& 61.07& 18.55 & 0& 19.44\\
        \hline
        $\tau=1$& 94.26& 72.53& 0& 19.19& 74.58& 46.47 & 0& 11.26& 62.93& 18.66 & 0& 19.09\\
        \hline
        $\tau=10$& 95.41& 73.94 &0.56& 39.79& 78.21& 50.67 & 0.29& 15.33& 64.70& 21.66 & 2.59& 23.88\\
        \hline
        $\tau=30$& 95.26& 74.93 & 91.09& 43.35& 78.27& 50.17 & 68.47& 18.81& 63.60& 21.30 & 49.45& 26.50\\
        \hline
        $\tau=50$& 94.92& 74.44 & 93.04& 36.13& 77.97& 49.87 &72.92& 20.50& 62.85& 20.40 & 54.95& 28.68\\
        \hline
        $\tau=70$& 95.05& 74.26 & 93.85& 35.43& 77.20& 49.61 & 73.49& 21.66& 62.14& 20.57 & 55.54& 30.14\\
        \hline
        $\tau=100$& 95.05& 73.08 & 94.29& 37.32& 77.14& 49.31 & 73.65& 22.83& 61.46& 18.82& 54.60& 32.71\\
        \bottomrule
    \end{tabular}
    }
    \label{tb:main_CNN}
\end{table*}

\begin{table}[t]
    \caption{Model performance and Robustness against Common Corruptions and Adversarial attacks (\%) under different temperatures with Transformer Vit-small-patch16-224. -C in the table represents the corresponding Common Corruptions and Perturbations set.}
    \centering
    \renewcommand{\arraystretch}{1.6}
    \setlength{\tabcolsep}{3pt}
    \scalebox{1}{
    \begin{tabular}{lcccccccc}
        \toprule
         \multirow{2}[2]{*}{Temp.}
         & \multicolumn{4}{c}{CIFAR10}& \multicolumn{4}{c}{CIFAR100}\\
        \cmidrule(lr){2-5} \cmidrule(lr){6-9} 
         & Clean& -C & PGD20& C\&W& Clean& -C& PGD20& C\&W\\
        \hline\hline
        $\tau=0.1$& 98.45& 92.83 & 0& 26.13& 89.79& 74.7& 0& 23.71\\
        \hline
        $\tau=0.5$& 98.33& 91.60 &0& 26.26& 90.53& 74.9 & 0& 29.25\\
        \hline
        $\tau=1$&  98.29& 92.21 & 0& 31.69& 90.78& 75.5 &0& 31.97\\
        \hline
        $\tau=10$&98.06& 92.19 &89.07& 31.89& 89.94& 75.5& 58.71& 34.96\\
        \hline
        $\tau=30$& 98.23& 91.72& 97.10& 38.21& 89.52& 74.6 & 86.25& 36.07\\
        \hline
        $\tau=50$& 98.22& 91.43 & 97.75& 39.52& 89.28& 73.8 & 87.29& 33.64\\
        \hline
        $\tau=70$& 98.03& 91.20& 97.72& 39.02& 89.48& 74.2 &87.96& 33.81\\
        \hline
        $\tau=100$& 98.07& 91.56&97.87& 38.26& 89.13& 73.47& 86.99& 31.84\\
        \bottomrule
    \end{tabular}}
    \label{tb:main_VIT}
\end{table}

\subsection{Training Convergence}

\begin{figure}[t]
    \centering
    \includegraphics[width=0.48\textwidth]{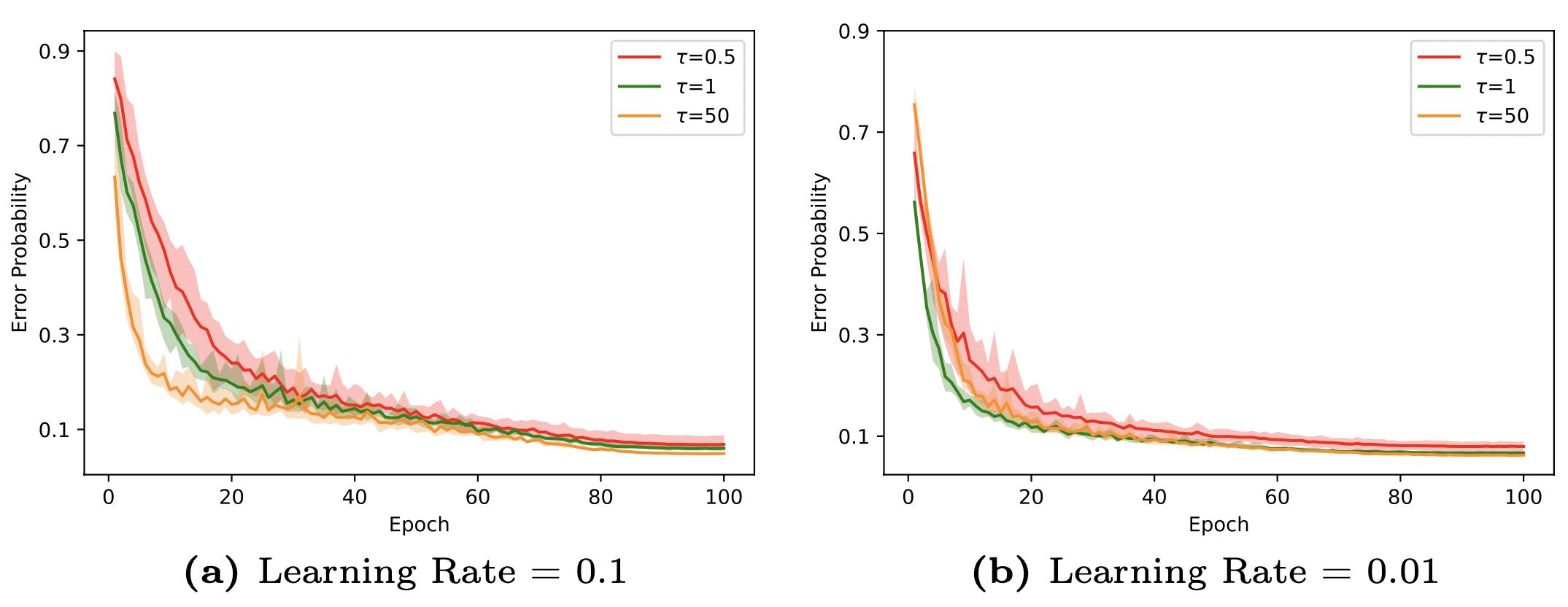}
    \caption{Test error number during training. The red line represents $\tau=0.5$, the green line represents $\tau=1$, and the orange line represents $\tau=50$. The model used is Resnet50 and is tested on CIFAR10. SGD optimizer is used during training with the learning rate set to 0.1 (a) and 0.01 (b). The shade areas consist of 6 total runs with different random seeds. The solid lines indicate the mean value across all runs.}
    \label{train-converge}
\end{figure}

We then conduct experiments observing the training process when applying different temperatures to the model. We validate the model on the test set every epoch and record the error probability. As our results shown in Fig.~\ref{train-converge}(a), we can clearly observe that not only does the training convergence speed increase as the temperature goes up, but models trained with higher temperatures also tend to converge to lower points, leading to better final performance. In fact, when we further decrease the temperature to around 0.1, the model would have a substantial risk of not converging at all. While this might appear contrary to the common understanding that focusing on hard classes will generally benefit the model, a more nuanced explanation is provided by delving further into the gradient analysis provided in Section 4.

From Eqn.~\ref{Eq:wi}, \ref{Eq:wj}, \ref{Eq:f}, we observe that if the logit of the target class is not the largest, its gradient will increase dramatically with low temperatures. This is potentially bad for models being known to converge inefficiently under large learning rates. One straightforward solution would be lowering the learning rate as shown in Fig.~\ref{train-converge}. While the training converging speeds are closer, the run with a higher temperature can still reach a better performance. Furthermore, regardless of the increase in overall training converging speed for $\tau=0.5$ and $\tau=1$ runs when lowering the learning rate, the final performances for all three runs actually get worse than runs with 0.1 learning rate. Therefore, this phenomenon cannot be attributed solely to a high learning rate. However, if we shift our perspective to the overall direction for optimization as done in Eqn.~\ref{Eq:batch_f}, it becomes clear that during the early stage of training, the encoder $f$ has not converged to an ideal point, leading to sub-optimal values produced for certain update directions. If this happens to be the direction of the target class and the \textit{error-prone class} which models with small temperatures tend to focus on, the model training can be impacted harmfully. In the meantime, high temperatures equalize the weight given to all the classes and ensure the update is not terribly wrong even if a few ${\partial L_{ce}(x)}/{\partial w_j}$ are in the wrong direction. Upon reaching this conclusion, we are surprised to find that this reasoning and our empirical observations align perfectly with the curriculum learning philosophy, that starting from hard samples may harm model optimization and learning outcomes.

\begin{figure*}[t]
    \centering
    \includegraphics[width=1\textwidth]{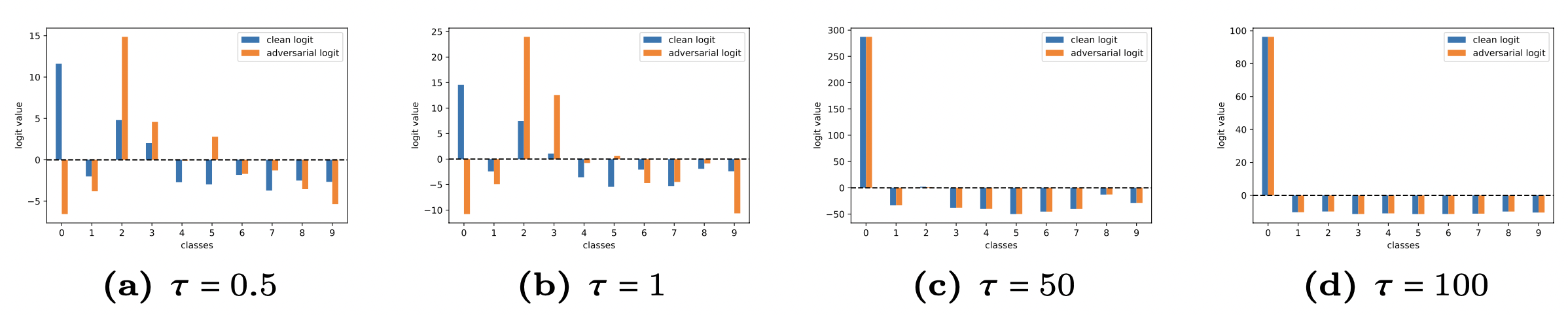}
    \caption{The logit changes before and after PGD20 attack. The blue lines stand for the logits of the samples before PGD attack, and the orange lines stand for the logits of the samples after PGD attack.}
    \label{logit-compare}
\end{figure*}

\subsection{Adversarial Robustness}

Table~\ref{tb:main_CNN} and Table~\ref{tb:main_VIT} show that models trained with elevated temperatures have strong adversarial robustness. TSNE plots in Fig.~\ref{cluster01} and Fig.~\ref{cluster02} also support this observation. This prompts questions regarding the mechanism behind the gained robustness. In this section, our focus is on investigating the model's behavior under adversarial attacks and understanding why the model demonstrates such robustness.

\textbf{Gradient analysis for adversarial generation.} In order to discern the source of model robustness, we follow the work in~\cite{Hou2023} and study the gradient of the classification loss with respect to the input to analyze the direction of the PGD attack, which can be written as
\begin{equation} \label{Eq:pgd}
    \begin{split}
        \frac{\partial L_{ce}}{\partial x} 
        =\begin{multlined}[t]
        [(\mathbb{S}(w^T_i\cdot f(x))-1)\cdot w^T_i+\\\sum_{j\neq i}w^T_j\cdot \mathbb{S}(w^T_j\cdot f(x)) ]
        \cdot \frac{\partial f(x)}{\partial x}
        \end{multlined}\\
    \end{split}
\end{equation}
As illustrated above, given a well-trained model, for most inputs where $\mathbb{S}(w^T_i\cdot f(x))	\approx1$, the gradient does not have a noticeable portion in target class $w_i$ on the early stage of the attack. This implies that rather than directly 'stepping away' from the target class, the attack will initially focus on approaching other class prototypes. Moreover, the second term, $\sum_{j\neq i}w^T_j\cdot \mathbb{S}(w^T_j\cdot f(x))$, indicates that all the other directions are weighted by their according probabilities. Therefore, untargeted attacks are actually targeted toward the \textit{error-prone class}, which most commonly is the largest probability class other than the target class. However, if a model lacks an \textit{error-prone class} given an input, all $w_k$ will be weighted equally. Consequently, the gradient would point toward all negative class prototypes, making it exceptionally challenging to determine the optimal direction. We noticed that such a scenario occurs when a model is trained with a small $\tau$. Then let's focus on the gradient update strength. For a data sample $x$ is classified correctly, $\mathbb{S}(w^T_j\cdot f(x))$ would be small when the model training temperature $\tau$ increases. That is, when a model is trained with high temperatures, not only the gradient direction to generate adversarial samples is not clear, but the gradient strength is also small. Both factors contribute to the robustness of the model when optimized with elevated temperatures.

\textbf{Raw Logit Analysis.} With the insight from the gradient analysis on adversarial attack, we then turn to observe the logit output around the adversarial attack, as shown in Fig.~\ref{logit-compare}. Each bar represents the logit value for each class, blue bars stand for the logit outputs of clean samples and orange bars are the logit outputs from adversarial samples. Models share similar characteristics in low temperatures, Fig.~\ref{logit-compare}(a,b), with the logit of the target class going down while the logit of the \textit{error-prone class} going up. However, for models trained with large temperatures, Fig.~\ref{logit-compare}(c,d), two logits are nearly identical with a minimal amount of changes. This contrasts the robustness gains during adversarial training, where the model learns the pattern of the adversarial noise.

\begin{figure*}[t]
    \centering
    
    \includegraphics[width=1\textwidth]{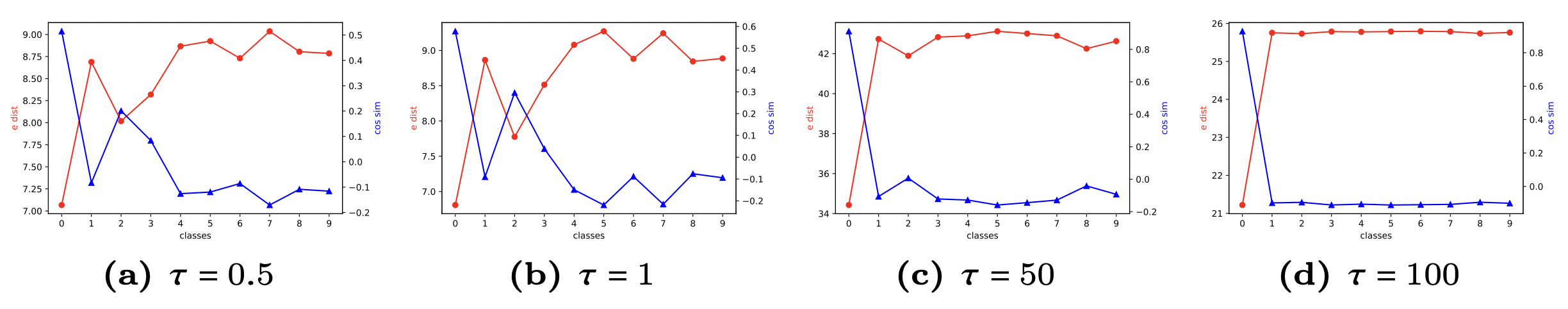}
        
    \caption{A demonstration of the Euclidean distance and cosine similarity between the encoded sample $f(x)$ and all class prototypes for one sample, with different temperature configurations. 
    The red lines indicate the Euclidean distance while the blue lines stand for cosine similarity.}
    \label{distance}
\end{figure*}

\textbf{Class Prototypes Analysis.} To further analyze the model behavior, we investigate the relation between the encoded feature, $f(x)$, and each class prototype, $w_j$. Here, we observe the Euclidean distance and cosine similarity. Fig.~\ref{distance} shows Euclidean distance and cosine similarity between one sample and all class prototypes. It is evident that as the training temperature goes up, the feature $f(x)$ tends to have an identical distance to all negative class prototypes. This indicates the model trained with high temperature is less likely to have an \textit{error-prone class}, which is essential for untargeted attacks as we discuss above. 

\begin{figure}[t]
    \centering
    \includegraphics[width=0.5\textwidth]{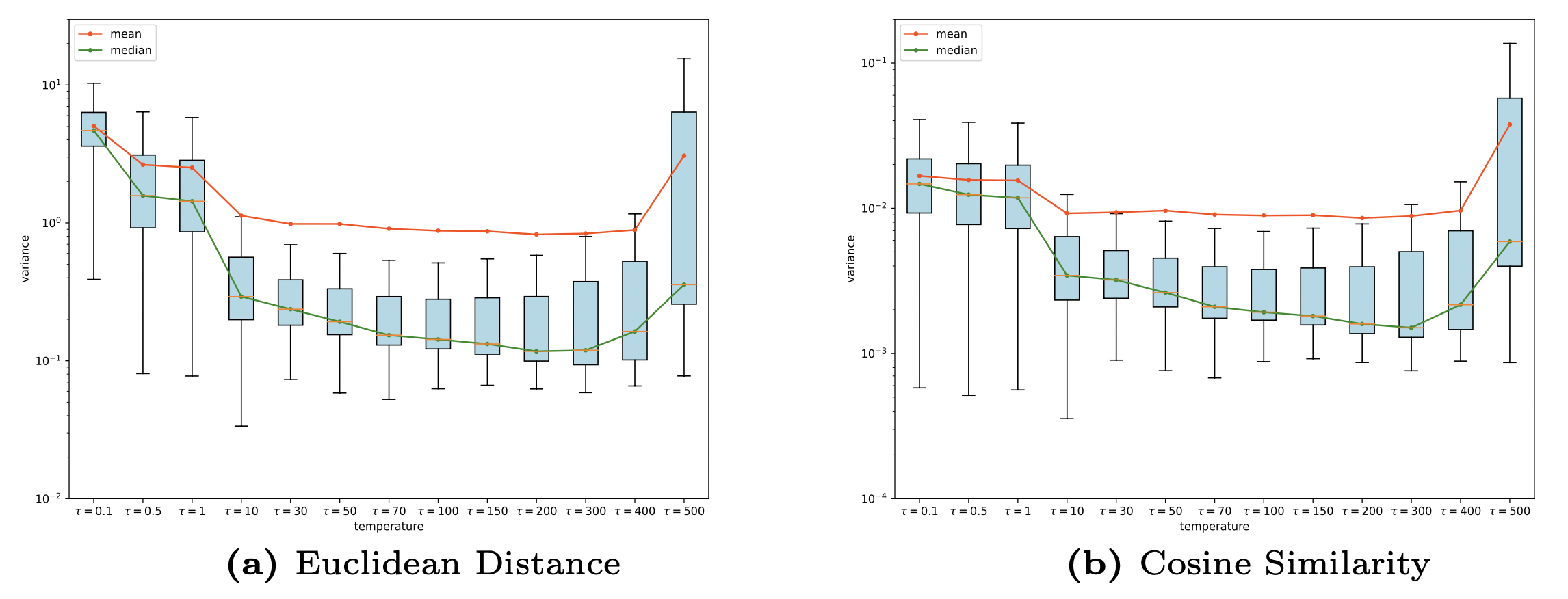}

    \caption{Box plot of the variance of the Euclidean distance and cosine similarity calculated from each sample. The variances are calculated across all negative class prototypes, therefore, lower variance indicates a more uniform distribution of all negative class distances. Each box is a model trained with a different temperature, the green line shows the median value across all variances and the orange line is the mean value of all variances.}
    \label{box_plot}
\end{figure}

Furthermore, to illustrate that the phenomenon shown in Fig.~\ref{distance} is not limited to one or a few samples, we calculate the variance of Euclidean distance and cosine similarity of all negative class prototypes across all samples in CIFAR10 test set. Note that as illustrated in Fig.~\ref{distance}, different models have very different ranges for Euclidean distance between encoded feature and class prototypes. Therefore, we map the value of different models into the same range to make a more direct comparison. Box plots are drawn in Fig.~\ref{box_plot} showing the overall variance results with each box being a model trained with a different temperature. We can observe a clear trend that when the temperature rises, the variance for both Euclidean distance and cosine similarity drops indicating the encoded sample, $f(x)$, has a more similar distance to all negative class prototypes. One might notice an increase in variance when the temperature reaches some threshold. We label them as extreme temperatures, which are so large that they can adversely affect the model's convergence.

\subsection{Further Discussion on Adversarial Robustness}
Despite the model trained with high temperatures showing superb robustness against untargeted PGD attack due to its nature attribute that discovers the weakness of PGD attack, it does not hold robustness against targeted attacks. The reason behind this is straightforward. In targeted attacks, Eqn.~\ref{Eq:pgd} no longer holds, and the gradient is not obligated to move towards all negative class prototypes with a weighted step size. Therefore, with the only source of the model robustness gained eliminated, it is naturally vulnerable to targeted attacks. 

\textbf{Remark}: Even though many attacks claim themselves to be untargeted attacks, they actually optimize toward one self-selected target, which we do not consider untargeted attacks under this setting. One popular example is the Difference of Logits Ratio(DLR) attack proposed by ~\cite{pmlr-v119-croce20b}. Regardless of its ability to rescale the logit,
\begin{equation}
    \mathrm{DLR}(x,y) = -\frac{z_y-\max\limits_{i\neq y} z_i}{z_{\pi1}-z_{\pi3}}
\end{equation}
shows that the DLR loss automatically selects the class holding the largest logit other than the target class as the attack target. Therefore, during optimization, it does not need to optimize toward all negative class prototypes. A similar example also includes FAB attack~\cite{pmlr-v119-croce20a}.

\subsection{Extended Experiment on Adversarial Training}
Given that our temperature control method is used inside the Cross-Entropy Loss, it is possible to apply this method in adversarial training. Here, we do preliminary experiments on the adversarial training baseline proposed by ~\cite{madry2018at} for the simplicity of its loss function. We add temperature control inside vanilla loss term forming
\begin{equation}
    L_{AT}(x,x_{adv},y,F)=L_{ce}(F(x)/\tau,y) + L_{ce}(F(x_{adv}),y),
\end{equation}
where $F$ is a combination of encoder and class prototypes.

\linespread{1.2}
\begin{table}[t]
    \setlength{\tabcolsep}{4pt}
    \centering
    \renewcommand{\arraystretch}{1.6}
    \caption{Preliminary experiments of adversarial training on CIFAR-10 with temperature control. The training scheme uses ~\cite{madry2018at} and the model is ResNet50.}
    \begin{tabular}{lccccccc}
        \toprule
        Temp.& $\tau=0.5$& $\tau=1$& $\tau=10$& $\tau=30$& $\tau=50$& $\tau=70$& $\tau=100$\\
        \hline
        Clean& 88.98& 85.67& 81.71& 82.62& 83.75& 84.28& 84.27\\
        \hline
        PGD20& 35.93& 42.63& 40.95& 44.96& 48.61& 49.16& 48.53\\
        \bottomrule
    \end{tabular}
    \label{tb:AT}
\end{table}
Our preliminary results are listed in Table~\ref{tb:AT}. We can clearly observe that model robustness increases as the temperature increases with a slight trade-off with clean accuracy, which confirms the possibility of combining the temperature control method with adversarial training. While further extension to other adversarial training methods is possible, it remains a complex problem for most adversarial training involves complex loss functions that may introduce terms other than the Cross-Entropy function. Also, balancing the vanilla loss term and adversarial loss term largely relies on empirical experiments. Therefore, further exploration of fitting this into other adversarial training methods falls beyond the scope of this paper. 

\section{Conclusion \& Limitation}
In this paper, we investigate the under-explored property of temperature scaling with the softmax function on image classification tasks. By performing gradient analysis with the Cross-Entropy classification loss and executing different empirical experiments, we show that temperature scaling can be a significant factor in model performance. Further experiments reveal applying high temperatures during training introduces enormous robustness against gradient-based untargeted adversarial attacks. We hope our work raises the interest of other researchers to utilize the simple temperature scaling in the common Cross-Entropy loss.

One limitation of this study was that we didn't report an explicit algorithm to set the best temperature values. We will work on this in our future work. One takehome note, as a hyperparameter, the tuning cost of the tempeerature is low as a wide range of temperatures (30 to 70) can provide improvements to the model.

{
    \small
    \bibliographystyle{splncs04}
    \bibliography{ref_xh}
}

\end{document}